\documentclass{opt2021} % Author names withheld

\usepackage{algorithm,algorithmic}
\usepackage{amssymb, multirow, paralist, color,amsmath}
\newtheorem{thm}{Theorem} 

\newtheorem{ass}{Assumption}
\usepackage{enumitem}
\usepackage{textcase} 
\usepackage[T1]{fontenc}
\usepackage[utf8]{inputenc} % allow utf-8 input
\usepackage{hyperref}       % hyperlinks
\usepackage{url}            % simple URL typesetting
\usepackage{booktabs}       % professional-quality tables
\usepackage{nicefrac}       % compact symbols for 1/2, etc.
\usepackage{microtype,graphics}      % microtypography
\usepackage{xcolor}         % colors
\usepackage{makecell}

\def \O {\mathcal{O}}

\def \R {\mathbb{R}}

\def \w {\mathbf{w}}
\def \v {\mathbf{v}}

\def \x {\mathbf{x}}

\def \E {\mathrm{E}}

\def \x {\mathbf{x}}

\def \e {\mathbf{e}}

\def \1 {\mathbf{1}}

\def \z {\mathbf{z}}
\def \s {\mathbf{s}}

\def \y {\mathbf{y}}
\def \u {\mathbf{u}}

\def \y {\mathbf{y}}
\def \E {\mathrm{E}}
\def \x {\mathbf{x}}

\def \z {\mathbf{z}}
\def \u {\mathbf{u}}

\def \w {\mathbf{w}}

\def \R {\mathbb{R}}

\def \v {\mathbf{v}}

\def \s {\mathbf{s}}

\def \teta {\tilde{\eta}} 
\newcommand{\required}[1]{\section*{\hfil #1\hfil}}

\title[A Novel Convergence Analysis for Algorithms of the Adam Family]{A Novel Convergence Analysis for Algorithms of the Adam Family}    

\optauthor{%
\Name{Zhishuai Guo$^\dagger$} \Email{zhishuai-guo@uiowa.edu}\\
\Name{Yi Xu$^\ddagger$} \Email{yixu@alibaba-inc.com}\\
\Name{Wotao Yin$^\ddagger$}
\Email{wotao.yin@alibaba-inc.com}\\
\Name{Rong Jin$^\ddagger$}
\Email{jinrong.jr@alibaba-inc.com}\\
\Name{Tianbao Yang$^\dagger$} 
\Email{tianbao-yang@uiowa.edu}\\
   \addr$^\dagger$Department of Computer Science, The University of Iowa, Iowa City, IA 52242, USA \\
       \addr$^\ddagger$Machine Intelligence Technology, Alibaba Group, Bellevue, WA 98004, USA \\
}

\begin{document}

\maketitle
 
\begin{abstract} 
Since its invention in 2014, the Adam optimizer~\citep{kingma2014adam} has received tremendous attention. On one hand, it has been widely used in deep learning and many variants have been proposed, while on the other hand their theoretical convergence property remains to be a mystery. It is far from satisfactory in the sense that some studies require strong assumptions about the updates, which are not necessarily applicable in practice, while other studies still follow the original problematic convergence analysis of Adam, which was shown to be not sufficient to ensure convergence.  Although rigorous convergence analysis exists for Adam, they impose specific requirements on the update of the adaptive step size, which are not generic enough to cover many other variants of Adam.  To address theses issues,  in this extended abstract, we present a simple and generic proof of convergence for a family of Adam-style  methods (including Adam, AMSGrad, Adabound, etc.). Our analysis only requires an increasing or large "momentum" parameter for the first-order moment, which is indeed the case used in practice,  and a boundness condition on the adaptive  factor of the step size, which applies to all variants of Adam under mild conditions of stochastic gradients.  We also establish a variance diminishing result for the used stochastic gradient estimators. Indeed, our analysis of Adam is so simple and generic that it can be leveraged to establish the convergence for solving a broader family of non-convex optimization problems, including min-max, compositional, and bilevel optimization problems. For the full (earlier) version of this extended abstract, please refer to~\cite{guo2021stochastic}.    %Our analysis is based on a widely used but not fully understood stochastic estimator using moving average (SEMA), which only requires a general unbiased stochastic oracle. %In particular, we analyze Adam-style methods based on the variance recursion property of SEMA for stochastic non-convex minimization. 
\end{abstract}
 
%\begin{keywords}%
%  List of keywords%
%\end{keywords} 

\section{Introduction} 
Stochastic adaptive methods originating from AdaGrad for convex minimization~\citep{duchi2011adaptive,mcmahan2010adaptive} have attracted tremendous attention for stochastic non-convex optimization~\citep{ada18bottou,ada18orabona,adagradmom18,tieleman2012lecture,chen2018closing,luo2018adaptive,zeiler2012adadelta}. Adam~\citep{kingma2014adam} is an important variant of AdaGrad, which is widely used in practice for training deep neural networks, and many variants of Adam were proposed for improving its performance, e.g.~\citep{DBLP:conf/nips/ZaheerRSKK18,luo2018adaptive,Liu2020On}.
Its analysis for non-convex optimization has also received a lot of attention~\citep{DBLP:conf/iclr/ChenLSH19}. 
For more generality, we consider a family of Adam-style algorithms. The update is given by 
 \vspace{-0.05in} 
\begin{equation}\label{eqn:adam}
\hspace*{-0.3in}\text{Adam-style:}\quad\left\{\begin{aligned}
&\v_{t+1} = (1-\beta_t)\v_t + \beta_t \O_F(\x_t),\\
&\u_{t+1} = h_t (\O_F(\x_0), \ldots, \O_F(\x_t)), \\
&\x_{t+1}  = \x_t - \eta_t \frac{\v_{t+1}}{\sqrt{\u_{t+1}} + G_0}, t=0, \ldots, T.
\end{aligned}\right. 
\end{equation}
where $h_t$ denotes an appropriate mapping function whose specific choices are given later. 

One criticism of Adam is that it might not converge for some problems with inappropriate  momentum parameters. In particular, the authors of AMSGrad~\citep{reddi2019convergence} show that Adam with small momentum parameters can diverge for some problems. However, we notice that the failure of Adam shown in AMSGrad~\citep{reddi2019convergence} and the practical success of Adam come from an inconsistent setting of the momentum parameter for the first-order moment. In practice, this momentum parameter (corresponding to $1-\beta_t$ in (\ref{eqn:adam})) is usually set to a large value (e.g., 0.9).  However, in the failure case analysis of Adam~\citep{reddi2019convergence} and many existing analysis of Adam and their variants~\citep{DBLP:conf/nips/ZaheerRSKK18,luo2018adaptive,Liu2020On,DBLP:conf/iclr/ChenLSH19,naichenrmsprop}, such momentum parameter is set as a small value or a decreasing sequence. We provide the first  analysis of Adam and other variants with a more natural increasing or large momentum parameter for the first-order moment. 

Several recent works have tried to prove the (non)-convergence of Adam. In particular, Zou et al. \cite{DBLP:conf/cvpr/ZouSJZL19} establish some sufficient condition for ensuring Adam to converge. In particular, they choose to increase the momentum parameter for the second-order moment and establish a convergence rate in the order of $\log(T)/\sqrt{T}$, which was similarly established in \cite{2020arXiv200302395D} with some improvement on the constant factor. Zaheer et al. \cite{DBLP:conf/nips/ZaheerRSKK18} show that Adam with a sufficiently large mini-batch size can converge to an accuracy level proportional to the inverse of the mini-batch size. Chen et al.~\cite{DBLP:conf/iclr/ChenLSH19} analyze the convergence properties for a family of Adam-style algorithms. However, their analysis requires a strong assumption of the updates to ensure the convergence, which does not necessarily hold as the authors give non-convergence examples.  Different from these works, we give an alternative way to ensure Adam converge by  using an increasing or large momentum parameter for the first-order moment without any restrictions on the momentum parameter for the second-order moment and without requiring a large mini-batch size. This seems more natural and consistent with the practice.  Indeed, our analysis is applicable to a family of Adam-style algorithms, and is agnostic to the method for updating the normalization factor in the adaptive step size as long as it can be upper bounded.  The large momentum parameter for the first-order moment is also the key part that differentiates our convergence analysis with existing non-convergence analysis of Adam~\cite{DBLP:conf/iclr/ChenLSH19,reddi2019convergence,kingma2014adam}, which require  the  momentum parameter for the first-order moment to be decreasing  to zero or sufficiently small. 

A key in the analysis is to carefully utilize the design of the stochastic estimator of the gradient. Traditional methods that simply use an unbiased gradient estimator of the objective function are not applicable to many problems and also suffer slow convergence due to large variance of the unbiased stochastic gradients.  Recent studies in stochastic non-convex optimization have proposed better stochastic estimators of the gradient based on variance reduction technique (e.g., SPIDER, SARAH, STORM)~\citep{fang2018spider,wang2019spiderboost,pham2020proxsarah,cutkosky2019momentum}. However, these estimators sacrifice generality as they require that the unbiased stochastic oracle is Lipschitz continuous with respect to the input, which prohibits many useful tricks in machine learning for improving {\bf generalization} and {\bf efficiency} (e.g., adding random noise to the stochastic gradient~\citep{DBLP:journals/corr/NeelakantanVLSK15},  gradient compression~\citep{NIPS2017_6c340f25,pmlr-v70-zhang17e,10.5555/3326943.3327063}). In addition, they also require computing stochastic gradients at two points per-iteration, making them further restrictive.

In Adam-style methods, the stochastic estimators are based on the moving average. 
%We first give a brief introduction of moving average estimator and then summarize our contributions.  
In order to generate a sequence of iterates $\{\x_0, \x_1, \ldots, \x_T \}$, we usually need to  track another sequence of $\{g(\x_0), g(\x_1), \ldots, g(\x_T)\}$, where $g$ is a Lipschitz continuous mapping that is useful for constructing the gradient of the objective function. However, $g(\x_t)$ can be only accessed through {\bf an unbiased stochastic oracle} denoted by $\O_g$ such that for any input $\x$, it returns a random variable $\O_g(\x)$ satisfying $\E[\O_g(\x)] = g(\x)$.  For more generality, we do not assume that $\O_g$ is Lipschitz continuous with respect to the input even $g$ is Lipschitz continuous. One example of such stochastic oracle is $O_g(\x) = g(\x; \zeta) + \xi$, where $\E_{\zeta}[g(\x; \zeta)]= g(\x)$ and  $\xi$ is a zero-mean random noise (e.g., zero-mean Gaussian noise). Therefore, the variance-reduced stochastic estimators based on SPIDER, SARAH or STORM are not applicable. Instead, we will consider another  stochastic estimator based on moving average, i.e., we maintain and update a sequence of $\{\z_1, \ldots, \z_T\}$ by 
\begin{align}\label{eqn:SEMA}
    \z_{t+1}  = (1-\beta_t)\z_t + \beta_t \O_g(\x_t), \quad t=0, \ldots, T.
\end{align}
We refer to this estimator sequence for tracking $\{g(\x_1), \ldots, g(\x_T)\}$ as stochastic moving average estimator (\textbf{SEMA}) in contrast to SPIDER/SARAH/STORM. It the literature, stochastic methods that employ the above estimator are usually referred to as momentum methods~\citep{cutkosky2020momentum,DBLP:conf/iclr/ChenLSH19,NEURIPS2020_d3f5d4de}, with $1-\beta_t$ called the "momentum" parameter. 
% We use the same tradition to name some of our newly proposed methods but refer to~(\ref{eqn:SEMA}) as SEMA specifically. 

Besides in Adam-style methods, SEMA has been widely used in other stochastic non-convex optimization methods such as in stochastic compositional minimization~\citep{YMSP,wang2017stochastic,ghadimi2020single,wang2017accelerating}. Although the SEMA has been widely used in practice, its power for solving a broad range of stochastic optimization problems has not been fully discovered.   We present a simple and intuitive proof of convergence of  a family of Adam-style algorithms with an increasing or large momentum parameter for the first-order moment, which include many variants such as Adam, AMSGrad, Adabound, AdaFom, etc. A surprising result is that Adam with an increasing or large momentum parameter for the first-order moment indeed converges at the same rate as SGD without any modifications  on the update  or restrictions on the "momentum" parameter for the second-order moment. In our analysis, $\beta_t$ is decreasing in the same order of step size, which yields an increasing "momentum" parameter $1-\beta_t$. This increasing momentum parameter is more natural than the decreasing (or small) momentum parameter, which is indeed the reason that makes Adam diverge on some examples~\citep{reddi2019convergence}. Our increasing/large "momentum" parameter $1-\beta_t$ is also consistent with that the large value close to 1 used in practice~\citep{kingma2014adam}. To the best of our knowledge, this is the first time that Adam was shown to converge for non-convex optimization with a more natural large "momentum" parameter for the first-order moment. 
We also prove that averaged variance of the stochastic estimator of the gradient decreases over time. A comparison of the results in this paper with existing results is summarized in Table~\ref{tab:0}. Moreover, our analysis can be extended to analyze the convergence of Adam-style algorithms for a broader family of non-convex optimization problems, including compositional optimization, min-max optimization and bilevel optimization problems~\cite{guo2021stochastic}. 

\begin{table*}[t] 
	\caption{Comparison with previous results. "mom. para." is short for momentum parameter. 1st and 2nd are short for first order and second order, respectively. "-" denotes no strict requirements and applicable to a range of updates. $\uparrow$ represents increasing as iterations and $\downarrow$ represents decreasing as iterations. $\epsilon$ denotes the target accuracy level for the objective gradient norm, i.e., $\E[\|\nabla F (\x)\|]\leq \epsilon$. 
}\label{tab:0} 
	\centering
	\label{tab:1}
	\scalebox{0.9}{\begin{tabular}{lllccc}
			\toprule
			Problem& Method	&batch size&\makecell{$\uparrow$ or $\downarrow$ \\ 1st mom. para.}&\makecell{$\uparrow$ or $\downarrow$\\ 2nd mom. para.}&Converge?\\
			\hline  
			&This work&$O(1)$&$\uparrow$&-&Yes \\
			&\citep{kingma2014adam}&$O(1)$&$\downarrow$&constant&No\\
	Non-convex		&\citep{DBLP:conf/iclr/ChenLSH19}&$O(1)$&Non-$\uparrow$&-&No\\
	(Adam-family)	&\citep{DBLP:conf/nips/ZaheerRSKK18}&$O(1/\epsilon^2)$&constant&$\uparrow$&Yes\\
			&\citep{DBLP:conf/cvpr/ZouSJZL19}&$O(1)$&constant&$\uparrow$&Yes\\
			&\citep{2020arXiv200302395D}&$O(1)$&constant&$\uparrow$&Yes\\ 
			\hline 
	\end{tabular}}
	\vspace*{-0.15in} 
\end{table*} 

\section{Notations and Preliminaries}
% 	\vspace*{-0.1in} 
{\bf Notations and Definitions.} Let $\|\cdot\|$ denote the Euclidean norm of a vector or the spectral norm of a matrix.   Let $\|\cdot\|_F$ denote the Frobenius norm of a  matrix. A mapping $h$ is $L$-Lipschitz continuous iff $\|h(\x) - h(\x')\|\leq L\|\x - \x'\|$ for any $\x, \x' \in \R^d$.  A function $F$ is called $L$-smooth if its gradient $\nabla F(\cdot)$ is $L$-Lipschitz continuous. A function $g$ is $\lambda$-strongly convex iff $g(\x)\geq g(\x') + \nabla g(\x')^{\top}(\x  - \x') + \frac{\lambda}{2}\| \x - \x'\|^2$ for any $\x, \x'$.   A function $g(\y)$ is called $\lambda$-strongly concave if $-g(\y)$ is $\lambda$-strongly convex.  For a differentiable function $f(\x, \y)$, we let $\nabla_x f(\x, \y)$ and $\nabla_y f(\x, \y)$ denote the partial gradients with respect to $\x$ and $\y$, respectively. Denote by $\nabla f(\x, \y) = (\nabla_x f(\x, \y)^{\top}, \nabla_y f(\x, \y)^{\top})^{\top}$. 
%Let $\Pi_{\Omega}$ denote a projection onto a convex set $\Omega$. With $\Omega_R=\{\x\in\R^d: \|\x\|\leq R\}$, we also use $\Pi_{R} = \Pi_{\Omega_R}$ for simplicity.
Let $\circ$ denote an element-wise product. We denote by $\x^2$, $\sqrt{\x}$  an element-wise square and element-wise square-root, respectively. 

We will consider non-convex minimization~(\ref{eqn:snc}), which has broad applications in machine learning. This paper focuses on theoretical analysis and our goal for these problems is to find an $\epsilon$-stationary solution of the primal objective function $F(\x)$ by using stochastic oracles. 
\begin{definition}
Consider a differentiable function $F(\x)$,  a randomized solution $\x$ is called an $\epsilon$-stationary point if it satisfies $\|\nabla F(\x)\| \leq \epsilon$. 
\end{definition} 
% Depending on the problem's structure, we require different stochastic oracles that will be described for each problem later. 
Before ending this section, we present the widely used stochastic momentum method for solving non-convex minimization problem $\min_{\x\in\R^d} F(\x)$ through an unbiased stochastic oracle that returns a random variable $\O_F(\x)$ for any $\x$ such that $\E[\O_F(\x)] = \nabla F(\x)$. For solving this problem, the stochastic momentum method (in particular stochastic heavy-ball (SHB) method) that employs the SEMA update is given by 
\begin{equation}\label{eqn:sma}
\hspace*{-0.3in}\quad\left\{ \begin{aligned}
&\v_{t+1} = (1-\beta)\v_t + \beta\O_F(\x_t)\\
&\x_{t+1} = \x_t - \eta \v_{t+1}, \quad t= 0, \ldots, T.
\end{aligned}\right.
\end{equation}
where $\v_0 = \O_F(\x_0)$. In the literature,  $1 - \beta$ is known as the momentum parameter and $\eta$ is known as the step size or learning rate. It is notable that the stochastic momentum method can be also written as $\z_{t+1} = \beta\z_t - \eta\O_F(\x_t)$, and $\x_{t+1} = \x_t + \z_t$~\cite{yangnonconvexmo}, which is equivalent to the above update with some parameter change shown in Appendix~\ref{appendix:SMM}. The above method has been analyzed in various studies~\cite{ghadimi2020single, NEURIPS2020_d3f5d4de,yu_linear,yangnonconvexmo}. Nevertheless, we will give a unified analysis for the Adam-family methods with a much more concise proof, which covers SHB as a special case. A core to the analysis is the use of a known variance recursion property of the SEMA estimator stated below. 
\begin{lemma}({\bf Variance Recursion of SEMA})[Lemma 2, \citep{wang2017stochastic}]
Consider a moving average sequence $\z_{t+1} = (1-\beta_t)\z_t + \beta_t \O_h(\x_t)$ for tracking $h(\x_t)$, where $\E_t[\O_h(\x_t)] = h(\x_t)$ and $h$ is a $L$-Lipschitz continuous mapping. Then we  have
\begin{align*}
\E_t[\|\z_{t+1} - h(\x_t)\|^2]\leq (1-\beta_t)\|\z_t - h(\x_{t-1})\|^2+ 2\beta_t^2\E_t[\|\O_h(\x_t) - h(\x_t)\|^2]  + \frac{L^2\|\x_{t} - \x_{t-1}\|^2}{\beta_t}.
\end{align*}
where $\E_t$ denotes the expectation conditioned on all randomness before $\O_h(\x_t)$.  
\label{lem:sema_mengdi}
\end{lemma}
We refer to the above property as variance recursion (VR) of the SEMA. 

%\begin{align}\label{eqn:snc}
%\min_{\x\in\R^d} F(\x),
%\end{align} 
%and
%\begin{align}\label{eqn:mm}
%\min_{\x\in\R^d}F(\x) = \max_{\y\in \Y}f(\x, \y). 
%\end{align}
%and
%\begin{equation}\label{eqn:sbo_formulation}
%\begin{split}
%&\min_{\x\in\R^d}F(\x) = f(\x, \y^*(\x))\\
%&s.t.~\y^*(\x)=\arg\min_{\y\in \Y}g(\x, \y). 
%\end{split} 
%\end{equation} 

\section{A Novel Analysis of Adam with a Large  Momentum Parameter} 
In this section, we consider the standard stochastic non-convex minimization, i.e., 
\begin{align}\label{eqn:snc}
\min_{\x\in\R^d} F(\x),
\end{align} 
where $F$ is smooth and is accessible only through an unbiased stochastic oracle. These conditions are summarized below for our presentation. 
\begin{ass}\label{ass:1}Regarding problem~(\ref{eqn:snc}), the following conditions hold: 
\vspace{-0.03in}  
\begin{itemize}[leftmargin=*]
\item $\nabla F$ is $L_F$ Lipschitz continuous; 
\vspace{-0.1in}
\item $F$  is accessible only through an unbiased stochastic oracle that returns a random variable $\O_F(\x)$ for any $\x$ such that $\E[\O_F(\x)] = \nabla F(\x)$, and $\O_F$ has a variance bounded by $\E[\|\O_F(\x)- \nabla F(\x)\|^2]\leq \sigma^2(1 + c\|\nabla F(\x)\|^2)$ for some $c>0$. 
\end{itemize} 
\end{ass}
{\bf Remark:} Note that the variance bounded condition is slightly weaker than the standard condition $\E[\|\O_F(\x)- \nabla F(\x)\|^2]\leq \sigma^2$. An example of a random oracle that satisfies our condition but not the standard condition is $\O_F(\x) = d\cdot\nabla F(\x)\circ\e_i$, where $i\in\{1,\ldots, d\}$ is randomly sampled and  $\e_i$ denotes the $i$-th canonical vector with only $i$-th element equal to one and others zero.  For this oracle, we can show that $\E[\O_F(\x)] = \nabla F(\x)$ and $\E[\|\O_F(\x) - \nabla F(\x)\|^2]\leq (d-1)\|\nabla F(\x)\|^2$.    

In the following we present a novel analysis of Adam-style methods based on VR of SEMA. The update rule of Adam-style rules has been given in \eqref{eqn:adam}. 
% For more generality, we first present a family of Adam-style algorithms. The update is given by 
% \vspace{-0.05in} 
%\begin{equation}\label{eqn:adam}
%\hspace*{-0.3in}\text{Adam-style:}\quad\left\{\begin{aligned}
%&\v_{t+1} = (1-\beta_t)\v_t + \beta_t \O_F(\x_t),\\
%&\u_{t+1} = h_t (\O_F(\x_0), \ldots, \O_F(\x_t)), \\
%&\x_{t+1}  = \x_t - \eta_t \frac{\v_{t+1}}{\sqrt{\u_{t+1}} + G_0}, t=0, \ldots, T.
%\end{aligned}\right. 
%\end{equation}
%where $h_t$ denotes an appropriate mapping function.  
A key to our convergence analysis of Adam-style algorithms is the boundness of the  step size scaling factor $\s_t = 1/(\sqrt{\u_{t+1}} + G_0)$, which is presented as an assumption below for more generality. We denote by $\teta_t = \eta_t \s_t$. 
 \vspace{-0.07in}
 \begin{ass}\label{ass:2}
 For the Adam-style algorithms in~(\ref{eqn:adam}), we assume that $\s_t =  1/(\sqrt{\u_{t+1}} + G_0)$ is upper bounded and lower bounded, i.e., there exists $0<c_l<c_u$ such that $c_l\leq \|\s_t\|_\infty\leq c_u$. 
 \end{ass}
 \vspace{-0.07in}  
 {\bf Remark:} Under the standard assumption $\|\O_F(\x)\|_\infty\leq G$~\citep{kingma2014adam,reddi2019convergence}, we can see many variants of Adam will satisfy the above condition.  Examples include Adam~\citep{kingma2014adam}, AMSGrad~\citep{reddi2019convergence}, AdaFom~\citep{DBLP:conf/iclr/ChenLSH19}, Adam$^+$~\citep{DBLP:journals/corr/abs-2011-11985},  whose $\u_t$ shown in Table~\ref{tab:2} all satisfy the above condition under the bounded stochastic oracle assumption. 
Even if the condition $\|\O_F(\x)\|_\infty\leq G$ is not satisfied, we can also use the clipping idea to make $\u_t$ bounded. This is used in Adabound~\citep{luo2018adaptive}, whose $\u_t$ is given by 
\vspace{-0.08in}
\begin{align*}
&\text{Adabound: } \u'_{t+1} = (1-\beta_t')\u'_t + \beta_t' \O_F^2(\x_t), \quad \u_{t} = \Pi_{[1/c_u^2, 1/c_l^2]}[\u_{t+1}],\quad G_0 =0,
\end{align*}  
% \vspace{-0.03in}  
where $c_l\leq c_u$ and $\Pi_{[a, b]}$ is a projection operator that projects the input into the range $[a, b]$. We summarize these updates and their satisfactions  of Assumption~\ref{ass:2} in Table~\ref{tab:2}. Note that SHB also satisfies Assumption \ref{ass:2} automatically. 

\begin{table*}[t] 
	\caption{Different Adam-style methods and their satisfactions of Assumption~\ref{ass:2}}
	\label{tab:2} 
	\centering
	% \label{tab:2}
	\scalebox{0.9}{\begin{tabular}{lccc}
			\toprule
		method&update for $h_t$&Additional assumption&$c_l$ and $c_u$ \\
		\midrule
				SHB &$\u_{t+1} = 1,G=0$ & - & $c_l =1, c_u =1$ \\
\midrule
		Adam& $\u_{t+1} = (1-\beta_t')\u_t + \beta_t' \O_F^2(\x_t)$&$\|\O_F\|\infty\leq G$&$c_l \geq \frac{1}{G+G_0}, c_u \leq \frac{1}{G_0}$\\
		\midrule
		AMSGrad& \makecell{$\u'_{t+1} = (1-\beta_t')\u'_t + \beta_t' \O_F^2(\x_t)$\\ $\u_{t+1} = \max(\u_t, \u'_{t+1})$}&$\|\O_F\|\infty\leq G$&$c_l \geq \frac{1}{G+G_0}, c_u \leq \frac{1}{G_0}$\\
		\midrule 
		\makecell{AdaFom \\(AdaGrad)}& $\u_{t+1} = \frac{1}{t+1}\sum_{i=0}^t\O^2_F(\x_i)$ &$\|\O_F\|\infty\leq G$&$c_l \geq \frac{1}{G+G_0}, c_u \leq \frac{1}{G_0}$\\ 
		\midrule
        Adam$^+$& $\u_{t+1} = \|\v_{t+1}\|$& $\|\O_F\|\leq G$ & $c_l \geq \frac{1}{\sqrt{G}+G_0}, c_u \leq \frac{1}{G_0}$\\  
        \midrule
        Adabound& \makecell{$\u'_{t+1} = (1-\beta_t')\u'_t + \beta_t' \O_F^2(\x_t)$\\ $\u_{t} = \Pi_{[1/c_u^2, 1/c_l^2]}[\u'_{t+1}],\quad G_0 =0$}& - &  $c_l = c_l, c_u =c_u$\\
		\bottomrule
	\end{tabular}}
	\vspace*{-0.15in} 
\end{table*}

To prove the convergence of the update~(\ref{eqn:adam}). We first present a key lemma. %similar to Lemma~\ref{lem:1} with coordinate-wise step size. 
\begin{lemma}\label{lem:3}
For $\x_{t+1} = \x_t- \teta_t \circ \v_{t+1}$ with $\eta_t c_l\leq \teta_{t,i}\leq \eta_t c_u$ and $\eta_t L_F\leq c_l/(2c_u^2)$, we have
\vspace{-0.03in}
\begin{align*} 
&F(\x_{t+1})  \leq F(\x_t) +   \frac{ \eta_t c_u}{2}\|\nabla F(\x_t) - \v_{t+1}\|^2- \frac{\eta_t c_l}{2}\|\nabla F(\x_t)\|^2  - \frac{\eta_t c_l}{4}\|\v_{t+1}\|^2. 
\end{align*}
\end{lemma}

With the above lemmas, we can establish the following convergence of Adam-style algorithms.  
\begin{thm}\label{thm:2}
Let $\Delta_t = \|\v_{t+1} - \nabla F(\x_t)\|^2$ and $ F(\x_0) - F_*\leq \Delta_F$ where $F_* = \min\limits_{\x} F(\x)$. Suppose Assumptions~\ref{ass:1} and~\ref{ass:2} hold. With $\beta_t=\beta \leq \frac{\epsilon^2c_l}{12\sigma^2c_u}$, $\eta_t=\eta\leq \min\{\frac{\beta\sqrt{c_l}}{2L_F\sqrt{c_u^3}}, \frac{1}{\sqrt{2} L_Fc_u},\frac{c_l}{2c_u^2L_F}\} $, $T\geq\max\{\frac{6\Delta_0c_u}{\beta\epsilon^2c_l}, \frac{12\Delta_F}{\eta\epsilon^2 c_l}\}$, we have 
\begin{align*}
\E\left[\frac{1}{T+1}\sum_{t=0}^T \|\nabla F(\x_t)\|^2\right]\leq \epsilon^2, \quad \E\left[\frac{1}{T+1}\sum_{t=0}^T\Delta_t\right]\leq 2\epsilon^2. 
\end{align*}
\end{thm}
{\bf Remark:} We can see that the Adam-style algorithms enjoy an oracle complexity of $O(1/\epsilon^4)$ for finding an $\epsilon$-stationary solution. To the best of our knowledge, this is the first time that the Adam with a large momentum parameter $1-\beta$ was proven to converge. One can also use a decreasing step size $\eta_t\propto 1/\sqrt{t}$ and decreasing $\beta_t =1/\sqrt{t}$ (i.e, increasing momentum parameter) and establish a rate of $\widetilde O(1/\sqrt{T})$ as stated below.  

\begin{thm}\label{thm:SHB_decrease} 
Let $\Delta_t = \|\v_{t+1} - \nabla F(\x_t)\|^2$ and $ F(\x_0) - F_*\leq \Delta_F$ where $F_* = \min\limits_{\x} F(\x)$. Suppose Assumption~\ref{ass:1} holds.   With 
 $c_1 = \min(1, \frac{1}{4c\sigma^2})$, 
$\beta_t=\frac{c_l}{8\sigma^2 c c_u \sqrt{t+1}}$, 
$\eta_t = \min\{\frac{\beta_t \sqrt{c_l}}{2 L_F \sqrt{c_u^3}}, \frac{1}{2L_F c_u}, \frac{c_l}{2c_u^2 L_F}\}$ and  $T\geq \widetilde{O}(\frac{c_u^5 c^2 L_F^2 \sigma^4 \Delta_F^2/c_l^5 + \Delta_0^2 c^2 c_u^4 \sigma^4/c_l^4 + 1/c}{\epsilon^4})$, we have  
\begin{align*}  
\E\left[\frac{1}{T+1}\sum_{t=0}^T \|\nabla F(\x_t)\|^2\right]\leq O(\epsilon^2), \quad  \E\left[\frac{1}{T+1}\sum_{t=0}^T\Delta_t\right]\leq O(\epsilon^2). 
\end{align*}
\end{thm}

\section{Conclusion \& Discussion}
In this paper, we have provided a simple and generic convergence analysis for a family of Adam-style methods for solving non-convex minimization problems.  We leveraged the variance recursion of the stochastic moving average estimator and established the convergence of practically used Adam and its variants. Our results bring some new insights to make the Adam method converge or convergence better. 

Indeed, the Lemma 3 paves the way for the convergence analysis of many Adam-style algorithms for solving a broader family of problems, including non-convex strongly concave min-max optimization problems, non-convex stochastic compositional optimization problems, and non-convex bilevel optimization problems. It is also worth mentioning that we can also prove a faster convergence of the Adam-style algorithms under a strong Polyak-\L ojasiewicz condition. We will explore this direction and examine how it will affect the convergence rate in the future work. 

Finally, it is worth mentioning that the oracle complexities established in this paper are optimal under a general stochastic unbiased oracle model. In addition, one can also replace the variance recursion of the stochastic moving average estimator by that of other stochastic estimators (e.g. STORM) to prove an optimal convergence under Lipchitz continuous oracle model.

\bibliography{ref,adam,minmax,recover,stagewise}
%\bibliography{ref,adam,minmax}
%%%%%%%%%%%%%%%%%%%%%%%%%%%%%%%%%%%%%%%%%%%%%%%%%%%%%%%%%%%% 

%%%%%%%%%%%%%%%%%%%%%%%%%%%%%%%%%%%%%%%%%%%%%%%%%%%%%%%%%%%%
\newpage
\required{ \hspace*{0.8in}\Large Appendix} 
\appendix 

\section{Stochastic Momentum Method}
\label{appendix:SMM}
In the literature~\cite{yangnonconvexmo}, the stochastic heavy-ball method is written as:
\begin{equation}\label{eqn:hb2}
\hspace*{-0.3in}\text{SHB:}\quad\left\{ \begin{aligned}
& \v'_{t+1} = \beta' \v'_{t} - \eta'\O_F(\x_t)\\
&\x_{t+1}  =  \x_t + \v_{t+1}, \quad t= 0, \ldots, T. 
\end{aligned}\right.
\end{equation}
To show the resemblance between the above update and the one in~(\ref{eqn:sma}), we can transform them into one sequence update:
\begin{align*}
&(\ref{eqn:sma}):  \x_{t+1} = \x_t - \eta \beta\O_F(\x_t) +   (1-\beta)(\x_t - \x_{t-1})\\
& \text{SHB: } \x_{t+1} = \x_t - \eta'\O_F(\x_t) + \beta' (\x_t - \x_{t-1}).
\end{align*}
We can see that SHB is equivalent to~(\ref{eqn:sma}) with $\eta'= \eta\beta$ and $\beta' =   (1-\beta)$. 

\section{Proof of Lemma~\ref{lem:3}}
\begin{proof}
%Let $\z_{t+1} = (\u_{t+1} -V_{t+1}[H_{t+1}]^{-1}\v_{t+1})$. 
Due to the smoothness of $F$, we can prove that under $\eta_t L_F\leq c_l/(2c_u^2)$
\begin{align*} 
&F(\x_{t+1}) \leq F(\x_t) + \nabla F(\x_t)^{\top} (\x_{t+1} - \x_t) + \frac{L_F}{2}\|\x_{t+1} - \x_t\|^2\\
&= F(\x_t) - \nabla F(\x_t)^{\top}(\teta_t\circ \v_{t+1}) + \frac{L_F}{2}\|\teta_t\circ\v_{t+1}\|^2\\ 
&  \leq F(\x_t) +   \frac{1}{2}\|\sqrt{\teta_t}\circ(\nabla F(\x_t) - \v_{t+1})\|^2- \frac{1}{2}\|\sqrt{\teta_t}\circ\nabla F(\x_t)\|^2 + (\frac{L_F}{2} \|\teta_t\circ\v_{t+1}\|^2 - \frac{1}{2}\|\sqrt{\teta_t}\circ\v_{t+1}\|^2 ) \\
&  \leq F(\x_t) +   \frac{\eta_t c_u}{2}\|\nabla F(\x_t) - \v_{t+1}\|^2- \frac{\eta_t c_l}{2}\|\nabla F(\x_t)\|^2   +  \frac{\eta_t^2 c_u^2L_F - \eta_t c_l}{2}\|\v_{t+1}\|^2\\ 
&  \leq F(\x_t) +   \frac{\eta_t c_u}{2}\|\nabla F(\x_t) -  \v_{t+1}\|^2- \frac{\eta_t c_l}{2}\|\nabla F(\x_t)\|^2  -  \frac{\eta_t c_l}{4}\|\v_{t+1}\|^2. 
\end{align*}
\end{proof}

\section{Proof of Theorem~\ref{thm:2}}
\begin{proof}
%First, we can see that $\eta_t = \eta/(\sqrt{\u_t} + G_0)\leq \eta[c_l, c_u]$, where $c_u = 1/G_0$ and $c_l = 1/(G+G_0)$. 
 By applying Lemma~\ref{lem:sema_mengdi} to $\v_{t+1}$, we have
\begin{align*}
\E_t[\Delta_{t+1}]\leq (1-\beta)\Delta_t+ 2\beta^2\sigma^2(1+c\|\nabla F(\x_{t+1})\|^2)  + \frac{L_F^2\|\x_{t+1} - \x_{t}\|^2}{\beta}.
\end{align*}
Hence we have, 
\begin{equation*}
\begin{split}
\E\left[\sum_{t=0}^T\Delta_t\right]\leq \E\left[\sum_{t=0}^T\frac{\Delta_t - \Delta_{t+1}}{\beta} + 2\beta\sigma^2(T+1) + 2\beta\sigma^2 c\sum_{t=0}^T\|\nabla F(\x_{t+1})\|^2+ \sum_{t=0}^T\frac{L_F^2\eta^2c_u^2\|\v_{t+1}\|^2}{\beta^2}\right].
\end{split}
\end{equation*}
 Adding the above inequality with Lemma~\ref{lem:3}, we have
\begin{align*}
&\frac{\eta c_l}{2}\E\left[\sum_{t=0}^T\|\nabla F(\x_t)\|^2\right] \leq F(\x_0) - F_* - \frac{\eta c_l}{4}\sum_{t=0}^T\|\v_{t+1}\|^2\\
&  + \frac{\eta c_u}{2}\E\left[\sum_{t=0}^T\frac{\Delta_t - \Delta_{t+1}}{\beta} + 2\beta\sigma^2(T+1) + 2\beta\sigma^2 c\sum_{t=0}^{T}\|\nabla F(\x_{t+1})\|^2+ \sum_{t=0}^T\frac{L_F^2\eta^2c_u^2\|\v_{t+1}\|^2}{\beta^2}\right]\\ 
&\leq   F(\w_0) - F_* - \frac{\eta c_l}{4}\sum_{t=0}^T\|\v_{t+1}\|^2 \\
&  + \frac{\eta c_u}{2}\E\left[\sum_{t=0}^T\frac{\Delta_t - \Delta_{t+1}}{\beta} + 2\beta\sigma^2(T+1) + 4\beta\sigma^2 c\sum_{t=0}^{T}\|\nabla F(\x_{t})\|^2 + 4\beta\sigma^2cL_F^2\eta^2c_u^2\|\v_{t+1}\|^2\right.\\
&\left.+ \sum_{t=0}^T\frac{L_F^2\eta^2c_u^2\|\v_{t+1}\|^2}{\beta^2}\right]
\end{align*}
Let $L_F^2\eta^2 c_u^3/(2\beta^2) \leq  c_l/8$ (i.e., $\eta\leq \frac{\beta\sqrt{c_l}}{2L_F\sqrt{c_u^3}})$ and $2\beta\sigma^2 c\leq c_l/(4c_u)$, $2\beta\sigma^2cL_F^2\eta^2c_u^3\leq c_l/8$ (i.e, $\eta L_F\leq \frac{1}{\sqrt{2}c_u}$), we have
\begin{equation*} 
\begin{split}
\frac{1}{T+1}\E\left[\sum_{t=0}^T\|\nabla F(\x_t)\|^2\right] &\leq \frac{\Delta_0c_u}{\beta T c_l}  +  \frac{2(F(\x_0) - F_*)}{\eta c_lT} + 2\beta\sigma^2\frac{c_u}{c_l}  +  \frac{1}{2}\frac{1}{T+1}\E\bigg[\sum_{t=0}^T\|\nabla F(\x_{t})\|^2\bigg].
\end{split}
\end{equation*}
As a result, 
\begin{align*}
\frac{1}{T+1}\E\left[\sum_{t=0}^T\|\nabla F(\x_t)\|^2\right] &\leq\frac{2\Delta_0c_u}{\beta T c_l}  +  \frac{4(F(\x_0) - F_*)}{\eta c_lT} + 4\beta\sigma^2\frac{c_u}{c_l}. 
\end{align*}
With $\beta \leq \frac{\epsilon^2c_l}{12\sigma^2c_u}$, $T\geq\max\{\frac{6\Delta_0c_u}{\beta\epsilon^2c_l}, \frac{12\Delta_F}{\eta\epsilon^2 c_l}\}$, we conclude the proof for the first part.  For the second part, 
we have
\begin{align*}
&\E\left[\sum_{t=0}^T\Delta_t\right]\leq \frac{\Delta_0}{\beta} + \beta\sigma^2(T+1) +\frac{c_l}{2c_u}\E\bigg[\sum_{t=0}^T\|\nabla F(\x_{t})\|^2\bigg] +  \E\bigg[\sum_{t=0}^T\frac{c_l}{2c_u}  \|\v_{t+1}\|^2\bigg]\\
& \leq \frac{\Delta_0}{\beta} + 2\beta\sigma^2(T+1) +\frac{1}{2}\E\bigg[\sum_{t=0}^T\|\nabla F(\x_{t})\|^2\bigg] +  \E\bigg[\sum_{t=0}^T\frac{1}{2}\Delta_t\bigg]
\end{align*}
As a result, 
\begin{align*}
&\E\left[\frac{1}{T+1}\sum_{t=0}^T\Delta_t\right]\leq \frac{2\Delta_0}{\beta T} + 4\beta\sigma^2 +\E\bigg[\frac{1}{T+1}\sum_{t=0}^T\|\nabla F(\x_{t})\|^2\bigg] \leq 2\epsilon^2. 
\end{align*}
\end{proof}
\section{Poof of Theorem~\ref{thm:SHB_decrease}} 
\begin{proof}
By applying Lemma \ref{lem:sema_mengdi} to $\v_{t+1}$, we have \begin{equation}      
\begin{split}
\E_t[\Delta_{t+1}] \leq (1-\beta_t) \Delta_t + 2\beta_t^2 \sigma^2 (1+c\|\nabla F(\x_{t+1})\|^2) + \frac{L_F^2\|\x_{t+1}-\x_t\|^2}{\beta_t}.
\end{split} 
\end{equation}
Hence we have
\begin{equation} 
\begin{split}
\E\left[ \sum\limits_{t=0}^T \beta_t \Delta_t\right] \leq \E\left[\sum\limits_{t=0}^{T}[\Delta_t - \Delta_{t+1}] + \sum\limits_{t=0}^{T} 2\beta_t^2\sigma^2(1+c\|\nabla F(\x_{t+1})\|^2) + \sum\limits_{t=0}^{T} \frac{L_F^2 \eta_t^2 c_u^2\|\v_{t+1}\|^2}{\beta_t} \right]. 
\end{split}
\end{equation}
Combining this with Lemma~\ref{lem:sema_mengdi}, 
\begin{equation}
\begin{split}
&\E\left[ \sum\limits_{t=0}^T \frac{\eta_t c_l}{2} \|\nabla F(\x_t)\|^2 \right] \\ 
&\leq \E\Bigg[ \sum\limits_{t=0}^{T} [F(\x_t) - F(\x_{t+1})] - \sum\limits_{t=0}^{T} \frac{\eta_t c_l}{4} \|\v_{t+1}\|^2  \\ 
& ~~~~~~~ + \frac{\eta_1 c_u}{2 \beta_1} \left[\sum\limits_{t=0}^T (\Delta_t - \Delta_{t+1}) + \sum\limits_{t=0}^{T} 2\beta_t^2 \sigma^2 (1+c\|\nabla F(\x_{t+1})\|^2) + \sum\limits_{t=0}^{T} \frac{L_F^2 \eta_t^2 c_u^2 \|\v_{t+1}\|^2}{\beta_t} \right] \Bigg] \\ 
&\leq \E\Bigg[ \sum\limits_{t=0}^{T} [F(\x_t) - F(\x_{t+1})] - \sum\limits_{t=0}^{T} \frac{\eta_t c_l}{4} \|\v_{t+1}\|^2 \\ 
& ~~~~~~~ + \frac{\eta_1 c_u}{2 \beta_1} \left[\sum\limits_{t=0}^T (\Delta_t - \Delta_{t+1})  +\sum\limits_{t=0}^{T} 2\beta_t^2 \sigma^2 +\sum\limits_{t=0}^{T} 4\beta_t^2 \sigma^2 c\|\nabla F(\x_{t})\|^2)  + \sum\limits_{t=0}^{T} 4\beta_t^2 \sigma^2 c L_F^2 \eta_t^2 c_u^2 \|\v_{t+1}\|^2 \right. \\
&~~~~~~~~~~~~~~~~~~~~~    \left. 
+ \sum\limits_{t=0}^{T} \frac{L_F^2 \eta_t^2 c_u^2 \|\v_{t+1}\|^2}{\beta_t} \right] \Bigg] \\ 
&\leq \E\left[F(\x_0) - F_* + \frac{\eta_1 c_u \Delta_0}{2\beta_1}  + \frac{2c_u \eta_1}{\beta_1 } \sum\limits_{t=0}^{T} \beta_t^2\sigma^2 + \sum\limits_{t=0}^{T}  \frac{\eta_t c_l}{4}\|\nabla F(\x_{t})\|^2) \right] ,
\end{split}  
\end{equation} 
where the last inequality holds because $\frac{2\eta_1 c_u}{\beta_1} \beta_t^2 \sigma^2 c \leq \frac{\eta_t c_l}{4}$, $\frac{2\eta_1}{\beta_1}\beta_t^2 \sigma^2 c L_F^2 \eta_t^2 c_u^3 \leq \frac{\eta_t c_l}{8}$ and $\frac{\eta_1}{2\beta_1} \frac{L_F^2 \eta_t^2 c_u^3}{\beta_t} \leq \frac{\eta_t c_l}{8}$. 
Hence,
\begin{equation}
\begin{split}
&\E[\sum\limits_{t=0}^{T} \eta_T c_l \|\nabla F(\x_t)\|^2] \leq \E[\sum\limits_{t=0}^{T} \eta_t c_l \|\nabla F(\x_t)\|^2] \\
&\leq \E\left[ 4(F(\x_0) - F_*) + \frac{2\eta_1 c_u \Delta_0}{\beta_1} + \sum\limits_{t=0}^T \frac{8 c_u \eta_1}{\beta_1} \beta_t^2 \sigma^2 \right] \\
&\leq \E\left[ 4(F(\x_0) - F_*) + \frac{\sqrt{c_l} \Delta_0}{L_F \sqrt{c_u}} + \sum\limits_{t=0}^T \frac{4 \sqrt{c_l}}{L_F \sqrt{c_u}} \beta_t^2 \sigma^2 \right] .
\end{split} 
\end{equation}
Thus,
\begin{equation}
\begin{split}
\E \left[ \frac{1}{T+1} \sum\limits_{t=0}^{T} \|\nabla F(\x_t)\|^2 \right] & \leq \E\left[ \frac{4(F(\x_0)-F(\x_*))}{\eta_T c_l (T+1)} + \frac{\Delta_0}{L_F \sqrt{c_l c_u} \eta_T (T+1)} +\sum\limits_{t=0}^{T} \frac{4\beta_t^2}{\eta_T L_F \sqrt{c_l c_u} (T+1)} \sigma^2 \right] \\
&\leq \frac{4\Delta_F}{\eta_T c_l (T+1)} +  \frac{\Delta_0}{L_F\sqrt{c_l c_u}\eta_T (T+1)} +  \frac{4 \sigma^2}{\eta_T L_F \sqrt{c_l c_u} (T+1)} \sum\limits_{t=0}^{T} \beta_t^2 \\ 
&\leq \frac{4\Delta_F}{\eta_T c_l (T+1)} + \frac{\Delta_0}{L_F \sqrt{c_l c_u} \eta_T (T+1)} + \frac{c_l^2}{16 \eta_T \sigma^2 L_F c^2 \sqrt{c_l c_u^5} (T+1)} \ln(T+2). 
\end{split} 
\end{equation}
Setting $T\geq \widetilde{O}(\frac{c_u^5 c^2 L_F^2 \sigma^4 \Delta_F^2/c_l^5 + \Delta_0^2 c^2 c_u^4 \sigma^4/c_l^4 + 1/c}{\epsilon^4})$, we conclude the proof for the first part.  
For the second part, we have 
\begin{equation} 
\begin{split}
\E\left[ \sum\limits_{t=0}^T \beta_t \Delta_t \right] \leq \Delta_0 + \frac{c_l^2}{16 \sigma^4 c^2 c_u^2} \ln(T+2) + \frac{1}{2} \E\left[  \sum\limits_{t=0}^{T} \beta_t \|\nabla F(\x_t)\|^2\right] 
+ \E\left[ \sum\limits_{t=0}^{T} \frac{1}{2} \beta_t \Delta_t \right] .
\end{split} 
\end{equation}
Then, 
\begin{equation}
\begin{split}
 \E\left[ \frac{1}{T+1} \sum\limits_{t=0}^{T} \Delta_t \right] \leq \frac{2\Delta_0}{\beta_T T} + \frac{c_l^2}{8 \sigma^4 c^2 c_u^2 (T+1)} \ln(T+2) 
 + \frac{1}{2} \E\left[\sum\limits_{t=0}^{T} \beta_t \|\nabla F(\x_t)\|^2\right],
\end{split}
\end{equation}
which concludes the proof of the second part. 
\end{proof} 
  
\end{document}